\documentclass[twoside]{article}
\usepackage{ecj,palatino,epsfig,latexsym,natbib}
\usepackage{graphicx,caption,afterpage,subfigure}
\usepackage{mathtools}
\usepackage{algorithm}
\usepackage{algorithmic}
\usepackage{array}
\usepackage{eqparbox}
\usepackage{color}

\begin{document}

\title{\bf Learning to Generate Genotypes with Neural Networks}  

\author{\name{\bf Alexander W. Churchill} \hfill \addr{alexanderchurchill@gmail.com}\\ 
        \addr{Queen Mary, University of London}
\AND
       \name{\bf Siddharth Sigtia} \hfill \addr{s.s.sigtia@qmul.ac.uk}\\
        \addr{Queen Mary, University of London}
\AND
       \name{\bf Chrisantha Fernando} \hfill \addr{chrisantha@google.com}\\
        \addr{Queen Mary, University of London}
}

\maketitle

\begin{abstract}

Neural networks and evolutionary computation have a rich intertwined history. They most commonly appear together when an evolutionary algorithm optimises the parameters and topology of a neural network for reinforcement learning problems, or when a neural network is applied as a surrogate fitness function to aid the evolutionary optimisation of expensive fitness functions. In this paper we take a different approach, asking the question of whether a neural network can be used to provide a mutation distribution for an evolutionary algorithm, and what advantages this approach may offer? Two modern neural network models are investigated, a Denoising Autoencoder modified to produce stochastic outputs and the Neural Autoregressive Distribution Estimator. Results show that the neural network approach to learning genotypes is able to solve many difficult discrete problems, such as MaxSat and HIFF, and regularly outperforms other evolutionary techniques.

\end{abstract}

\section{Introduction}
Evolutionary Algorithms typically employ static exploration methods, such as recombination and mutation, in order to traverse a search space. A problem with this approach is that ``building blocks", by which we mean structural features of the search space, can be easily broken, discarding important information acquired during the search process. Addressing this problem are Estimation of Distribution Algorithms (EDAs), which attempt to statistically model the current search space in order to uncover underlying structure and guide search towards optimal solutions in an efficient manner (\cite{pelikan2006scalable}).

At the heart of any EDA lies a model-building process. Examples include Bayesian Networks, Markov Networks and K-Means clustering (\cite{pelikan2002survey}). In this paper we introduce and investigate two neural-based methods for modelling: a Denoising Autoencoder (dA) and the Neural Autoregressive Distribution Estimator (NADE). An autoencoder is a feed forward neural network, consisting of at least one hidden layer, which is trained to reproduce its inputs from its outputs. Over the course of training, the hidden layer forms a, typically compressed, representation of the inputs (\cite{hinton2006reducing}). We explore two special types of autoencoder for learning effective genotypes. The dA is an autoencoder trained using inputs that are stochastically corrupted, a process that acts as a strong regulariser and also widens the basins of attraction of the outputs towards seen solutions. Although the dA is not a generative model, the encoding process can capture relationships between variables. When combined with stochastic outputs the dA can produce mutations to existing solutions, guided by the learnt encoded structure of the search space. We present a novel evolutionary algorithm, in which a model learns solution structure online from promising individuals in the population, and offspring are produced by propagating parents through the model.    

We also investigate the NADE, an autoencoder instance that explicitly learns a generative model from data, and has been shown to rival a Restricted Boltzmann Machine (RBM) but at a considerably lower computational cost (\cite{larochelle2011neural}). Results show that an evolutionary algorithm incorporating the autoencoder models is able to outperform a canonical genetic algorithm (GA) across a range of combinatorial optimisation problems, as well as the PBIL and BOA EDA methods in several cases.

\section{Background}

Since their introduction in \cite{holland1975adaptation}, evolutionary algorithms have themselves evolved, with extensive work exploring representation schemes, operators, parallelisation, co-evolution and the simultaneous optmisation of multiple objectives. A key concern has been to reduce the sensitivity to initial parameters, e.g. self-adaptive algorithms (\cite{back1993overview}), and automate the perceived dark art of designing genetic operators. Tackling the latter are Estimation of Distribution Algorithms (EDAs), also known as Probabilistic-Model-Building Genetic Algorithms, which are population based optimisers that typically replace genetic operators with a statistical model. The rationale behind the model building and related linkage learning approach (e.g. \cite{thierens2010linkage,harik1999linkage}) is that dependencies between variables can be captured and preserved. Search can be biased by sampling from the learnt structure of promising areas of the search space. Early work on EDAs concentrated on methods that explicitly modelled the probabilities of features occurring independently in a population of genotypes. These include the Compact Genetic Algorithm (\cite{harik1999compact}), the Population-Based Incremental Learner (PBIL) (\cite{baluja1994population}) and the Univariate Marginal Probability methods (\cite{pelikan2002survey}). More complex problems were able to be solved efficiently by modelling multivariate dependencies using clustering algorithms, e.g. the Extended Compact Genetic Algorithm (\cite{harik1999linkage}); Bayesian networks, e.g \cite{pelikan2002scalability}'s Bayesian Optimisation Algorithm (BOA); Markov networks and tree structures, among many others (\cite{pelikan2002survey}). The popular CMA-ES optimiser can be considered both a type of EDA and self-adaptive algorithm (\cite{hansen2003reducing}). 

Artificial neural networks and evolutionary computation have been combined together by many researchers. Evolutionary algorithms have been applied to train neural networks. In early research the synaptic weights of a neural network were directly encoded, such as in \cite{montana1989training}. Modern work, such as Hyper-NEAT (\cite{stanley2009hypercube}), employs indirect encodings that allow both the weights and topologies of large networks to be successfully optimised simultaneously. There has also been interest in employing neural networks directly in the search process. Early work showed that including learning in the evolutionary process can improve the effectiveness and efficiency of optimisation, e.g. (\cite{hinton1987learning,nolfi1994learning}). This is the standpoint taken in this paper but coming from a different perspective. The evolutionary algorithms in (\cite{hinton1987learning,nolfi1994learning}) are applied to optimising a neural network and integrate learning through backpropagation (\cite{rumelhart1988learning}) as a Baldwinian or Lamarckian process. Here, we use the unsupervised learning capabilities of neural networks to learn to generate incrementally more effective genotypes, either through a guided mutation (with the denoising autoencoder) or a learnt distribution (with the NADE).

With the `neural networks revival' of the last decade, which has seen neural network methods frequently outperforming other machine learning techniques, there has been interest in neural-based methods in an EDA context, particularly for multi-objective optimisation. A Growing Neural Gas (GNG) was used as a model in \cite{marti2008introducing}, employing a competitive Hebbian learning rule to cluster data without having to pre specify the number of groups. A shallow Restricted Boltzmann Machine (RBM) was used to model high dimensional data in (\cite{tang2010restricted,shim2013multi}), beating the state-of-the-art on several multi-objective continuous benchmark problems. Helmholtz machines have also been applied to this task with promising results (\cite{zhang2000bayesian}). Autoencoders (often referred to in the literature as auto-associators) are also neural-based method for unsupervised learning and have hitherto not been applied to combinatorial optimisation problems. Autoencoders have several interesting properties which can be favourably exploited to learn to generate genotypes in an evolutionary algorithm, which are discussed in more detail in the next section.

Another motivation for using an autoencoder approach is to investigate methods in which evolutionary algorithms can be implemented using neural structures. The {\em{Neural Replicator Hypothesis}} (\cite{fernando2010neuronal}) proposes that evolutionary dynamics could operate in the brain at ontogenetic timescales. In \cite{fernando2010neuronal}, a (1+1)-ES is implemented in a network of spiking neurons. Adding Hebbian learning enabled linkages to be found between features, and the 128-bit HIFF problem to be solved significantly faster than by a simple Genetic Algorithm. However, Hebbian learning is restricted to learning only pairwise relationships between variables, while autoencoders have the potential to learn complex multivariate interactions. Although the training method employed in this paper is not thought to be biologically plausible, success with these models would suggest that unsupervised learning in neural structures can provide an effective means of evolutionary search.

\section{Methods}

The main motivation for this paper is to assess whether neural network-based unsupervised learning methods can be used to learn good genotypes. In the evolutionary computation literature this topic has mainly been investigated in the context of EDAs. Typically, EDAs use unsupervised machine learning techniques to estimate distributions, i.e. build generative models, over the space of good quality solutions. Novel solutions are produced at each generation by updating and sampling from these distributions. An alternative approach is to use models that learn an encoding or compress the solution space, e.g. the generative grammar of \cite{cox2014solving}. Our proposed use of the denoising autoencoder deviates from the more standard generative model approach. Instead of learning a distribution over the entire solution space, we directly learn a mapping from a given input to a distribution over potential solutions. By carefully choosing the inputs to the autoencoder and the corruption noise, we can generate better solutions at each iteration. The autoencoder is a deterministic, feed-forward neural network architecture that does not presume a graphical structure like Bayesian networks or RBMs. The autoencoder can be trained online with mini-batch gradient descent which allows us to update the model at each iteration without having to rebuild the model every time. This implies that the network has a slowly decaying memory over the good solutions and that interesting features learnt at one generation can be carried forward to future generations.

We also explore the NADE for model building, which is a distribution estimator for high-dimensional binary data. The potential benefits of using the NADE are that it breaks down the difficult problem of learning a complex high-dimensional distribution to the problem of learning a product of one-dimensional distributions. The NADE is a tractable approximation to the RBM and is therefore computationally more efficient to train and sample from when compared to the RBM. As the RBM has already been applied successfully to multi-objective optimisation, it would be interesting to see how the NADE performs in an EDA-like context. The gradients of the objective function for training the NADE can be obtained exactly and therefore we can use powerful optimisers for better online training. The NADE can be trained online, meaning that information can be retained between iterations. Another advantage of using the NADE is that it can be easily extended to real-valued data. This is in contrast to other Bayesian EDAs like BOA. 



\subsection{The Autoencoder}


A standard autoencoder consists of an encoder and a decoder network. The encoder performs an affine transformation followed by an element-wise non-linear operation. The mapping performed by the encoder is deterministic and can be described as: $$h_{\theta}(\mathbf{x}) = f(\mathbf{Wx + b})$$ The decoder is also a deterministic network and operates on the encoded input $h_{\theta}(x)$ to produce the reconstructed input: $$ r_{\theta'}(\mathbf{h}) = g(\mathbf{W'h + b'})$$ The outputs of the encoder network can be interpreted as the encoding or representation that the autoencoder learns. The encoder network can be viewed as a non-linear generalisation of Principle Component Analysis (\cite{hinton2006reducing}). The autoencoder learns interactions between the input attributes and maps them to the hidden layer via a non-linear transformation, making it a powerful model for learning and exploiting the structure present in the best individuals.

Rather than using the outputs of the decoder network directly, we use them as parameters for a conditional distribution $p(X|Z=\mathbf{z})$ over the outputs of the network given an input $\mathbf{z}$. For binary optimization problems, the outputs $\mathbf{z}$ are considered to be parameters for the distribution $X|\mathbf{z} \sim \mathcal{B(\mathbf{z}})$ where $\mathcal{B}$ is the bernouilli distribution. For continuous parameter-optimisation problems, the outputs $\mathbf z$ parameterise a multi-variate normal distribution $X\ \sim\ \mathcal{N}(\mathbf{z},\,\sigma^2)$, where the covariance matrix is assumed to be diagonal and the standard deviation along the diagonal is a tuneable parameter.

The autoencoder learns an encoding or representation of the input space in order to recreate the inputs with high accuracy. In order to force the autoencoder to learn interesting representations, typically some kind of information bottleneck is applied to the representations. This bottleneck can either be enforced by reducing the dimensionality of the representation as compared to the input, or enforcing the hidden units to be sparse. A geometric interpretation of the representations learnt by an autoencoder is that it identifies the non-linear manifold on which the data is concentrated. The motivation for using an autoencoder is that it can discover structure, or alternatively the manifold on which the good solutions lie. Novel solutions can then be generated by exploring the manifold or sampling from the distribution learnt by an autoencoder. Due to the non-linear nature of the mapping learnt, it is possible to discover complex dependencies and structure in the data and therefore learn a more effective mutation distribution. 




In our algorithm we use a denoising autoencoder which is a variant of the standard autoencoder (\cite{vincent2008extracting}). The dA tries to recreate the input $\mathbf x$ from `corrupted' or `noisy' versions of the input, which are generated by a stochastic corruption process $\mathbf{\tilde x = q(\tilde x|x)}$. Adding the denoising criterion to the model forces the autoencoder to learn more robust representations which are invariant to small perturbations of the input. Alternatively, the denoising criterion can be seen as learning an operator that maps points that are far from the manifold towards points on or near the manifold on which the good solutions are concentrated. Although denoising was introduced in order to encourage the learning of more robust representations, in our system we employ the denoising criterion to widen the basins of attraction around the best individuals in every generation. During training, we train the autoencoder on the most promising solutions, and by increasing corruption we encourage individuals that are far away from the training set to move towards the nearest high quality solution. By making use of the corruption noise as a tuneable parameter, the extent of the basins of attraction can be controlled, with high corruption giving rise to large basins of attraction.

\subsection{Neural Autoregressive Distribution Estimator}

Although the outputs of the decoder network can be interpretted as the parameters of a probability distribution, the autoencoder is inherently a deterministic unsupervised learning algorithm. The Neural Autoregressive Distribution Estimator (NADE) is a bayesian network for density estimation of high dimensional binary data. The NADE can be viewed as an autoencoder architecture devised such that the output units represent valid probabilities. The initial inspiration for the NADE was as a tractable approximation to the Restricted Boltzmann Machine (RBM). Restricted Boltzmann Machines have been used in EDAs for multi-objective optimisation in the past (\cite{tang2010restricted}). However, the NADE has been demonstrated to be an accurate density estimator on its own and has been used for various applications (refs).

The NADE belongs to the family of fully visible Bayesian networks that factorise the joint distribution of some $D$ dimensional data point $x$ as a product of one-dimensional conditional distributions:
$$ P(x) = \prod_{i=1}^{D} P(x_i|x_{<i})$$ where $x_{<i} \equiv \left\{ x_k, \forall k < i \right\} $  and $D$ is the dimensionality of the input space. The main advantage of the factorisation above is that instead of trying to model a complex high-dimensional multi-modal distribution for $x$, the problem can be simplified to estimating the one-dimensional conditionals $P(x_i|x_{<i})$. A complex distribution can be estimated accurately if each conditional is learnt correctly. For the NADE: 
$$P(v_i=1|v_{i<i}) =  \sigma(b_i + W_{i,.}h_i)$$
$$ h_i = \sigma(c+W_{.,<i}x_{<i})$$ where $W_{.,<i}$ denotes a sub-matrix of $W$ with columns $1$ to $i-1$. 
The above network is equivalent to a feed-forward neural network for each conditional, with weight sharing between the neural networks. The tied connections distinguish the NADE from other fully visible Bayesian networks and make it computationally tractable to train and obtain probabilities at test time. Another advantage while using the NADE is that the log-likelihood and therefore the gradients can be computed exactly. This implies that the model can be trained with powerful gradient based optimisers without having to resort to sampling to estimate the log-likelihoods, as RBMs do. 

Both the NADE and the dA were trained using minibatch stochastic gradient descent (SGD). Here the gradients are calculated over a batch of inputs and the model parameters are updated with respect to these gradients. This differs from training where updates are made after each training example or after going through the entire training set. Minibatch SGD allowed us to parallelise training on a GPU while allowing the model to be updated online at each generation without rebuilding the model from random initialisation each time. 
We used a fixed learning rate and a batch size of 20 for all our experiments.

\subsection{Optimisation Algorithm}

The pipeline of the optimisation algorithm presented here is similar to canonical EDAs and is inspired by methods in HBOA (\cite{hboa}). Pseudocode is provided in Algorithm 1 below. A population of \(P\) solutions is maintained and updated at each iteration. An initial population of solutions is drawn from a uniform distribution. These solutions are evaluated and the fittest unique x\% are selected (i.e. using truncation selection) to be in the training set. The model is then trained with the training set for \(E\) epochs. Following training, \(P\) solutions are stochastically sampled from the model and stored in a sample set. Solutions from the sample set are then incorporated into the population.

There are slight differences in the optimisation process based on the neural network model. For the denoising autoencoder (GA-dA), the training set consists of the fittest unique x\% of solutions from the population. For generating samples in GA-dA, a new set of solutions is selected from \(P\) using tournament selection (with replacement) at each iteration to form an input set. Each member of the input set is propagated through the dA model, and the output vector, \(y\), is sampled from using a binomial distribution, to produce a new solution. For the NADE model (GA-NADE), solutions are sampled from the model until the sample set contains a set of solutions not present in the population, i.e. the sample set $ \cap $ the population. Both GA-dA and GA-NADE can use a niching technique. Although any method could be used, here we employ Restricted Tournament Selection (\cite{hboa}). A solution from the sample set is included in the population if it is better than its closest neighbour in a randomly chosen subset of the population of size, \(W\). Whether to use niching or not, as well as other parameters are determined through a grid search on a problem-by-problem basis.
 
\begin{algorithm}                      
\caption{Evolutionary optimisation with the NADE or denoising autoencoder models }          
\label{alg1}                           
\begin{algorithmic}                    
    \STATE $P \Leftarrow \text{population size}$ 
    \STATE $T \Leftarrow \text{\% population used for training}$
    \STATE $E \Leftarrow \text{no. epochs of backprop}$
    \STATE $LR \Leftarrow \text{learning rate for backprop}$
    \STATE $H \Leftarrow \text{no. hidden neurons}$
    \STATE $NICHING \Leftarrow \text{[True \OR False]}$\COMMENT{Use niching?}
    \STATE $W \Leftarrow \text{window size for niching}$
    \STATE $EVALS \Leftarrow \text{Max no. evaluations}$
    \STATE $evalsCount \Leftarrow \text{0}$
    \STATE $model \Leftarrow \text{dA \OR NADE}$ \COMMENT{Choose neural network model}
    \STATE $pop \Leftarrow \text{Initialise population}$
    \WHILE{$evalsCount < EVALS$}
        \STATE $trainingData = getTrainingData(model,pop,T)$
        \STATE $model = trainModel(model,trainingData)$
        \IF{$model$ is dA}
            \STATE $input = tournamentSelection(pop,P)$
            \STATE $samples = sampleModel(model,P,input)$
        \ELSE
            \STATE $samples = sampleModel(model,P)$
        \ENDIF
        \STATE $evaluate(samples) = i$
        \STATE $evalsCount = evalsCount + P$
        \IF{$NICHING$ is True}
            \STATE $pop = newPopulationNiching(pop,samples)$
        \ELSE
            \STATE $pop = samples$
        \ENDIF
    \ENDWHILE
    \RETURN $pop$
\end{algorithmic}
\end{algorithm}
\section{Experiments}
\label{sec:experiments}
The neural network-based optimisation algorithms (GA-dA and GA-NADE) are tested on several difficult discrete problems and their performance is compared to a canonical GA, PBIL and BOA. Below we describe the problems and problem instances tested and the comparison algorithms.
\subsection{Problems}
\label{sec:problems}
\subsubsection{Multi-dimensional Knapsack Problem}
Here we wish to choose of subset of \(N\) items to maximise the total value of the items, \(z\), while satisfying \(m\) constraints. We wish to maximise:
\vspace{2mm}\\
\(z = \sum_{j=1}^{N} v_jx_j\), subject to \(\sum_{j=1}^{N} w_{ij}x_i \leq c_i, i = 1, ..., m\)
\vspace{2mm}\\
\(x_i \in \{0,1\}, j = 1, ..., N\).
\vspace{2mm}\\
Two different instances are used in the results presented below. The first is the Weing8 instance (\cite{weingartner1967methods}), which has 105 items and two constraints (optimal solution is 602,319), and the second is a randomly generated instance with 500 items and one constraint (optimal solution is 100,104). If any of the \(m\) constraints are violated, fitness is the negative of the summed violations.
\subsubsection{Hierarchical If and only If}
The Hierarchical If and only If (HIFF) problem was created by Watson et al., as an example of a function which is not separable (\cite{watson1999hierarchically}). It is a pathological function for a hill climber because of a very large number of local optima. The HIFF is structured as a balanced binary tree (\cite{hboa}), where at each lower level a transfer function is applied to consecutive non-overlapping 2-bit partitions, 00 \(\rightarrow\) 0, 11 \(\rightarrow\) 1, anything else \(\rightarrow\) null, to define the string in the layer above. At each level, 00 and 11 blocks contribute to the fitness by \(2^{level}\). Two global optima exist, a string of all 1s and all 0s. In the results section, the algorithms are applied to the 128-bit and 256-bit problem instances. In order to remove any possible bias towards an all 1s solution, solutions from the neural network models have a random bit mask (fixed before each trial) applied to them before evaluating fitness.
\subsubsection{Royal Road}
The Royal Road function as defined by \cite{mitchell1992royal}, divides a bit string, \(x\), into a number of equally sized, non-overlapping partitions, \(z_i\in[x_i, ..., x_{i+n}])\). If all of the bits in the partition match a predefined pattern, \(s_i\) , then the partition's fitness contribution is added to the overall fitness. The existence of these ``Building blocks" was meant to act as a ``royal road" for GAs compared to hill-climbers but bad alleles hitchhiking to good partial solutions slows down convergence speed. The 128-bit problem with 16 8-bit partitions is defined as,
\\
\[
f(x)= \sum_{i=1}^{16} \delta_i(x)o(s_i), \text{\ \ where\ } \delta_i(x) \begin{dcases*}
1, \text{\ \ \ \ if } x\in s_i\\
0, \text{\ \ \ \ otherwise}
\end{dcases*}
\]
If a string contains two instances of s, a fitness of 16 is achieved. If all instances are found, fitness = 128. \(s_i\) instances are blocks of all 1s but as with the HIFF, a random mask is applied to solutions from the neural network models before fitness evaluation.
\subsubsection{Maximum Satisfaction}
The maximum satisfaction problem (MaxSat) is defined as finding an interpretation of predicates that satisfies the maximum number of clauses in a given predicate logical formula in conjunctive-normal form (CNF). This is known to be a difficult problem for GAs due to partial solutions leading away from the optimal solution (\cite{rana1998genetic,pelikan2003hierarchical}). Here, algorithms are tested on a 3-CNF, 100-bit MaxSat problem obtained from SATLIB. The problem consists of 430
clauses and belongs to the phase transition region, which is the point at which
the problem transitions from generally solvable to generally unsolvable.\subsection{Comparison Algorithms}
\label{sec:comparisons}
The neural networks are compared to a Genetic Algorithm (GA), Population-based Incremental Learning Algorithm (PBIL) and the Bayesian Optimisation Algorithm (BOA). The GA was chosen to see whether GA-dA and GA-NADE outperform a canonical evolutionary algorithm and thus tests whether the method has potential. The GA uses two-point crossover and bit-flip mutation.

PBIL is a univariate estimation of distribution algorithm which operates on a single genetic string, a probability vector that represents the likelihood of each variable being present in the population. Each element in the probability vector, \(p^n\),  is initially set to 0.5. The algorithm learns online, updating \(p\) using,
\[
p_i = (1-\alpha)p_i + \alpha b_i \ \ \ \ \forall{i \in{[1 ... n]}}
\]
where \(b\) is the best solution found in a set of samples from \(p\) at each iteration. At each iteration there is also a small chance of a mutation to each component of $p$. The algorithm was chosen to see how a univariate EDA compares to GA-dA and GA-NADE. Additionally, the two neural network methods are similar to PBIL as they also learn online.

BOA probabilistically models solutions present in a population by building a Bayesian network. At each iteration, a new Bayesian network, \(BN\), is constructed from a selection of solutions from the population selected by truncation selection. The joint distribution encoded in the network is then sampled from, and the samples are incorporated into the population. BOA was selected for comparison because the model captures multi-variate dependencies and the NADE model approximates a Bayesian network. In BOA the Bayesian network is constructed using a greedy algorithm, that restricts each node to having two parents at most, to limit computational cost.
\section{Results}
 \begin{table}[t!]
   \scalebox{0.8}{
    \begin{tabular}{ | p{1.8cm} | l | l | l | l | l | l | l | l | p{1cm} |}
    \hline
    Experiment & Algorithm & Min & Max & Mean & Mean Evals & Success \%\\ \hline
Royal Road 128 & GA & 120.0 & 128.0 & 127.20 \(\pm\)2 & 42600.00 \(\pm\)27034 & 90\%\\
 & dA & 128.0 & 128.0 & 128.00 \(\pm\)0 & 26500.00 \(\pm\)3721 & 100\%\\
 & NADE & 128.0 & 128.0 & 128.00 \(\pm\)0 & 50900.00 \(\pm\)9771 & 100\%\\
 & PBIL & 112.0 & 128.0 & 123.20 \(\pm\)6 & 74110.00 \(\pm\)23463 & 60\%\\
 & BOA & 128.0 & 128.0 & 128.00 \(\pm\)0 & 38175.00 \(\pm\)3118 & 100\%\\\hline
MaxSat & GA & 424.0 & 429.0 & 426.50 \(\pm\)1 & 500000.00 \(\pm\)0 & 0\%\\
 & dA & 429.0 & 430.0 & 429.78 \(\pm\)0 & 291944.44 \(\pm\)134967 & 70\%\\
 & NADE & 430.0 & 430.0 & 430.00 \(\pm\)0 & 141600.00 \(\pm\)28182 & 100\%\\
 & PBIL & 425.0 & 430.0 & 427.10 \(\pm\)1 & 492800.00 \(\pm\)35400 & 10\%\\
 & BOA & 427.0 & 429.0 & 428.30 \(\pm\)0 & 500000.00 \(\pm\)0 & 0\%\\\hline
HIFF 128 & GA & 832.0 & 1024.0 & 940.80 \(\pm\)86 & 416000.00 \(\pm\)87429 & 50\%\\
 & dA & 1024.0 & 1024.0 & 1024.00 \(\pm\)0 & 231000.00 \(\pm\)23537 & 100\%\\
 & NADE & 1024.0 & 1024.0 & 1024.00 \(\pm\)0 & 65500.00 \(\pm\)17492 & 100\%\\
 & PBIL & 560.0 & 772.0 & 652.00 \(\pm\)76 & 500000.00 \(\pm\)0 & 0\%\\
 & BOA & 1024.0 & 1024.0 & 1024.00 \(\pm\)0 & 29400.00 \(\pm\)1854 & 100\%\\\hline
HIFF 256 & GA & 1664.0 & 2304.0 & 1984.00 \(\pm\)225 & 1590500.00 \(\pm\)626719 & 30\%\\
 & dA & 2304.0 & 2304.0 & 2304.00 \(\pm\)0 & 1355500.00 \(\pm\)190793 & 100\%\\
 & NADE & 2304.0 & 2304.0 & 2304.00 \(\pm\)0 & 318333.33 \(\pm\)82663 & 90\%\\
 & PBIL & 1354.0 & 1544.0 & 1457.40 \(\pm\)48 & 2000000.00 \(\pm\)0 & 0\%\\
 & BOA & 2304.0 & 2304.0 & 2304.00 \(\pm\)0 & 98800.00 \(\pm\)7493 & 100\%\\\hline
Knapsack 500 & GA & 10047.0 & 10093.0 & 10074.20 \(\pm\)13 & 200000.00 \(\pm\)0 & 0\%\\
 & dA & 10096.0 & 10104.0 & 10102.70 \(\pm\)2 & 121620.00 \(\pm\)52114 & 70\%\\
 & NADE & 10051.0 & 10075.0 & 10060.33 \(\pm\)10 & 200000.00 \(\pm\)0 & 0\%\\
 & PBIL & 9984.0 & 10083.0 & 10039.00 \(\pm\)31 & 200000.00 \(\pm\)0 & 0\%\\
 & BOA & 10039.0 & 10099.0 & 10079.40 \(\pm\)17 & 200000.00 \(\pm\)0 & 0\%\\\hline
Weing8 Knapsack & GA & 619568.0 & 621086.0 & 620137.40 \(\pm\)511 & 100000.00 \(\pm\)0 & 0\%\\
 & dA & 621086.0 & 624319.0 & 623615.40 \(\pm\)1270 & 85400.00 \(\pm\)19172 & 60\%\\
 & NADE & 619886.0 & 624319.0 & 621483.50 \(\pm\)1479 & 91550.00 \(\pm\)17632 & 20\%\\
 & PBIL & 620060.0 & 624319.0 & 621304.80 \(\pm\)1566 & 97220.00 \(\pm\)5600 & 20\%\\
 & BOA & 578923.0 & 620821.0 & 599630.20 \(\pm\)17700 & 100000.00 \(\pm\)0 & 0\%\\\hline
    \end{tabular}
    }
    \caption{Results for the different search methods on 6 discrete problems. Showing the value of the minimum and maximum solution returned at the end of search, the mean best solution in the population after the end of search, the mean number of evaluations required to reach the optimum solution (or until the maximum number of evaluations) and the success rate for reaching the optimum, averaged across 10 trials.}
        \label{table:main_results}

\end{table}

The neural network methods and the three comparison algorithms described in Section \ref{sec:comparisons} were tested on the discrete problems described above. Before comparison, a grid search was performed on the parameters of all five algorithms, on a problem-by-problem basis. The parameter configuration that achieved the best fitness (averaged over three trials) was selected for comparison. On each problem, the algorithms were assessed over 10 independent trials.

\begin{figure}[t!]
\centering
\subfigure[MaxSat]{
\makebox[0.5\textwidth]{
        \includegraphics[scale=0.34]{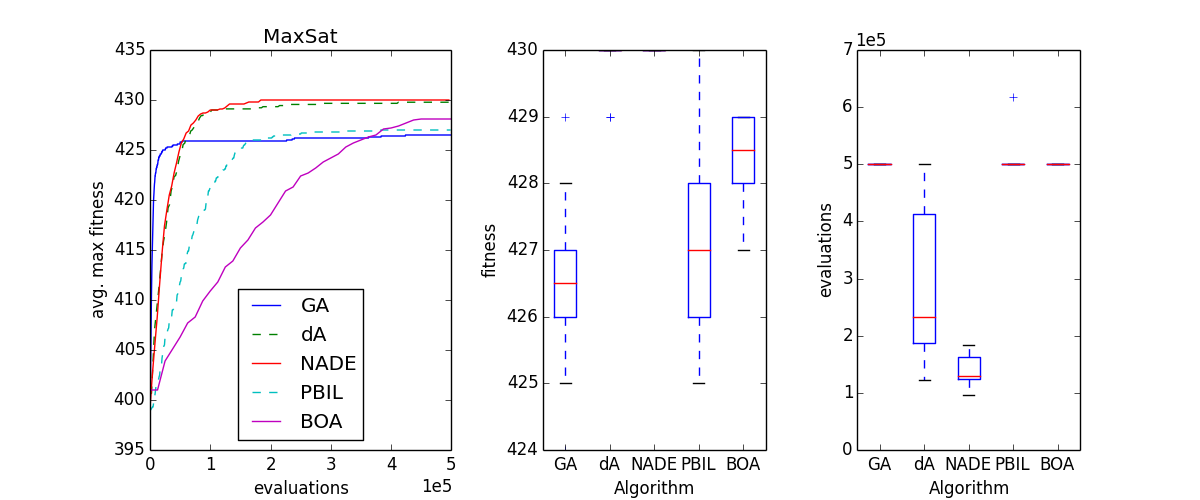}
    }
}
\subfigure[Knapsack (500 items, 1 constraint)]{
\makebox[0.5\textwidth]{
        \includegraphics[scale=0.34]{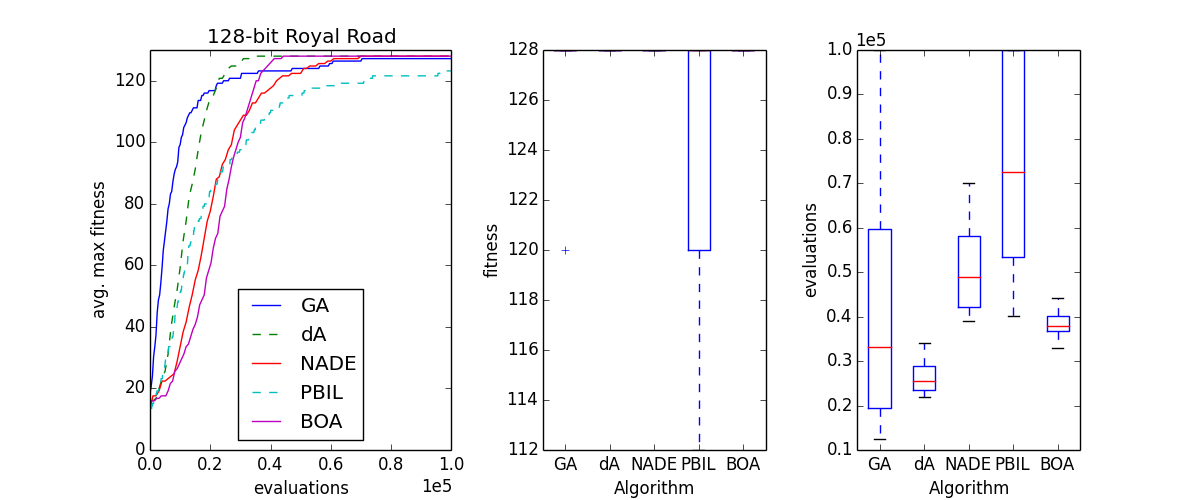}
    }
}
\subfigure[Weing Knapsack (105 items, 2 constraints)]{
\makebox[0.5\textwidth]{
        \includegraphics[scale=0.34]{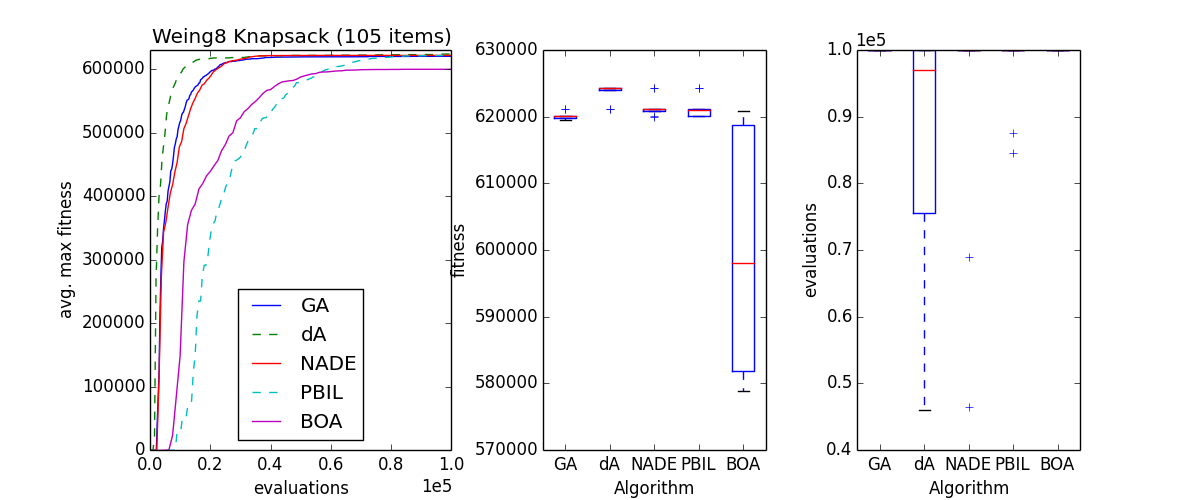}
    }
}
\caption[Optional caption for list of figures]{Showing results obtained by the different algorithms on MaxSat and Knapsack problems.}
\label{figure:results_plots1}
\end{figure}

\begin{figure}[t!]
\centering
\subfigure[500-item knapsack (1 constraint)]{
\makebox[0.5\textwidth]{
        \includegraphics[scale=0.34]{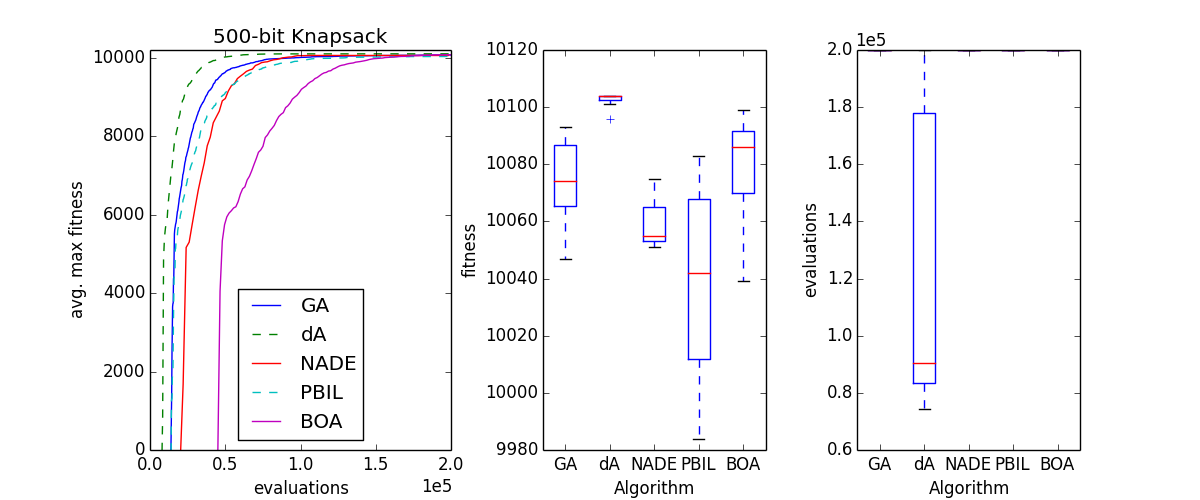}
    }
}
\subfigure[128-bit HIFF]{
\makebox[0.5\textwidth]{
        \includegraphics[scale=0.34]{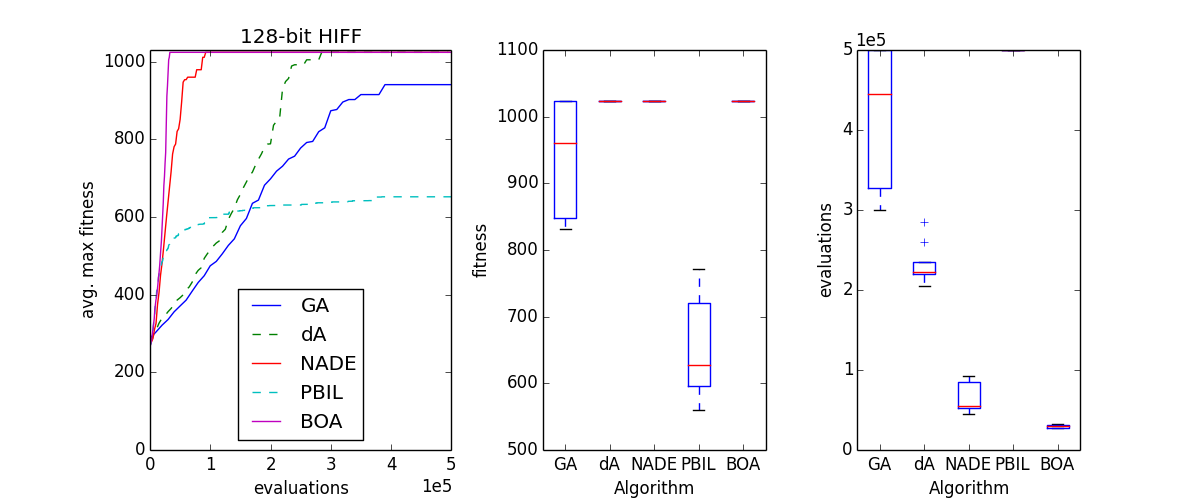}
    }
}
\subfigure[256-bit HIFF]{
\makebox[0.5\textwidth]{
        \includegraphics[scale=0.34]{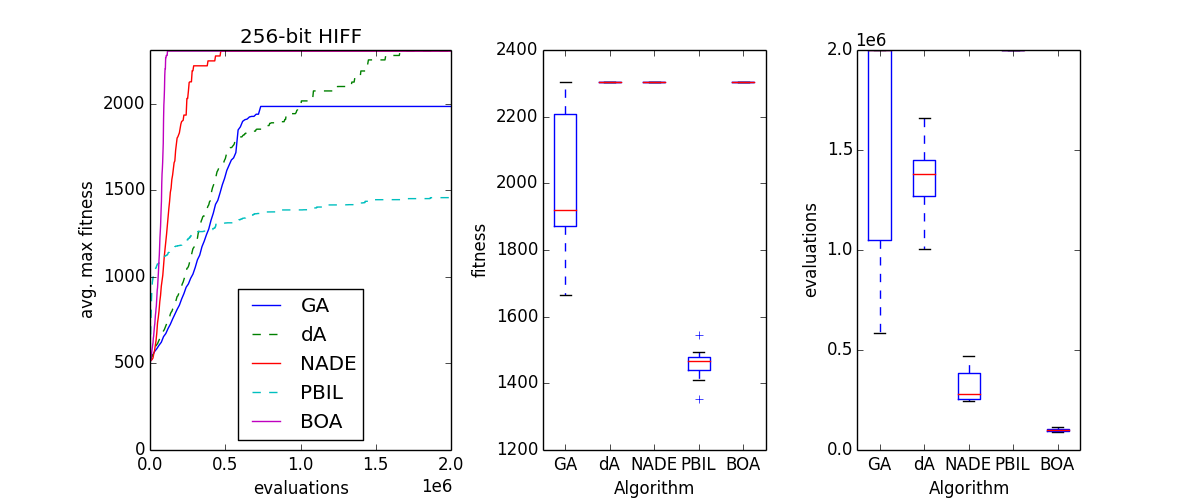}
    }
}
\caption[Optional caption for list of figures]{Showing results obtained by the different algorithms on Royal Road and HIFF problems.}
\label{figure:results_plots2}
\end{figure}

Figures \ref{figure:results_plots1} and \ref{figure:results_plots2} show plots of the best found fitness as the number of evaluations increase, averaged across ten independent trials, as well as boxplots showing the distribution of the best found solutions at the end of the optimisation process, for each of the six test problems. Results are also summarised in table \ref{table:main_results}. We observe that on four of the six problems, either GA-dA or GA-NADE performs the best, with BOA outperforming both on the two HIFF problems.

On the MaxSat problem, GA-NADE is the only algorithm that reaches the optimal solution on all 10 trials. Although both GA-NADE and GA-dA progress in a similar manner earlier on in the search process, the NADE method continues to improve and finds the optimal solution of 430 satisfied clauses much more quickly, within around 140,000 evaluations. Interestingly BOA is unable to find the optimal solution within the evaluation limit, although it performs much better than PBIL and the GA.

GA-dA is the clear winner on both knapsack problems. On the 500-item instance it is the only algorithm that locates the optimal solution and on the Weing8 instance (which has two constraints) it has a far greater success rate than the next best performer. On both problems it also improves its best solution much more quickly than the other algorithms. Both GA-NADE and BOA have much slower rates of improvement than GA-dA, although GA-NADE is considerably faster than BOA. BOA finds slightly better solutions than GA-NADE on the 500-item knapsack but performs considerably worse on the Weing8 instance.

On both tested HIFF instances BOA is the far better performer, solving the 128-bit problem twice as fast and the 256-bit instance three times faster than GA-NADE. In turn, GA-NADE is able to solve the 128-bit instance in half the evaluations required by GA-dA and a quarter as many on the 256-bit instance. GA-dA is better than the standard GA on this problem, solving it faster and more consistently. The GA is not able to solve the HIFF instances on every trial, although it can locate the optimal solution while PBIL cannot.

GA-dA displays the best performance on the Royal Road problem, consistently solving it within a small number of evaluations. The next best is the GA which can solve it more quickly but produces a much greater variance in discovery time. BOA is able to solve the problem much faster on average than GA-NADE, with low variance in terms of the number of evaluations taken to solve the problem.

\section{Discussion}

\subsection{Algorithmic Comparison}
The results presented in the preceding section demonstrate that the denoising autoencoder (GA-dA) and NADE (GA-NADE) methods perform well across a number of difficult discrete problems. GA-dA outperforms the other methods on both of the knapsack problems as well as the Royal Road, and GA-NADE is the only algorithm that is able to consistently solve the MaxSat problem instance. GA-dA outperforms the GA on all test problems in terms of the quality of the final solution and/or the number of evaluations required. GA-NADE outperforms the GA on all of the problems apart from the Royal Road and the 500-item knapsack where they are not significantly different. BOA outperforms both neural network methods on the HIFF problems. It also outperforms GA-NADE on the Royal Road and Knapsack-500 problems but not on the others. This suggests that BOA is better able to discover the type of dependancy present in the HIFF problem. GA-NADE performs considerably better than GA-dA, which suggests that the Bayesian network-based generative model is well-suited to discovering the structure in the search space present in this problem. The poor performance of PBIL on the HIFF problem indicates that a model capable of learning complex multivariate dependencies is needed. The structure learnt by GA-dA allows it to significantly outperform the GA but it is much slower than BOA or GA-NADE on this problem. 

A MaxSat problem is difficult because there are a large number of interdependencies between variables and solving subproblems in the form of individual clauses can lead you away from the optimal solution \citep{rana1998genetic}. GA-NADE is able to solve this problem consistently while BOA is not. This problem could be well suited to the NADE model which assumes a dependency between every variable - each variable is either a parent or a child of another. Thus a complex joint distribution of the variables is learnt. The dA is also able to learn multi-variate dependency with the potential advantage of having no precedence constraints. Both GA-dA and GA-NADE approach the optimal solution in far fewer evaluations than BOA, although it takes GA-dA many more iterations to reach the optimal solution of 430 clauses, from the sub-optimal state of 429.

Knapsack problems are interesting for three reasons, they have have real-world applications in resource management, they do not have obvious linkages and they are constrained. A 3-CNF MaxSat problem is complex because individual variables are present in multiple clauses and changing one variable can have a large impact on the fitness of the sequence. Due to the nature of the constraint handling of these knapsack problems, a small change in the genotype can have a huge impact on fitness. If a solution changes from being under to over capacity, its value will swing from positive to negative. If there were no constraints, a knapsack problem would reduce to a MaxOnes problem and all variables would be fully independent. Variables can still be treated independently, which effectively will assign a higher probability to high value or denser items. Given more constraints the dependencies between groups of variables become more complicated as certain combinations of items are required in order to satisfy the constraints. In terms of the rate of improvement, PBIL performs much better on the single constraint knapsack problem but takes much longer to get close to the optimal solution compared to the other algorithms. BOA has a much slower improvement rate on the 2-constraint Weing8 instance and does not find as good solutions. Interestingly, on this problem the GA uses a high crossover rate of 0.5 and thus swaps a large number of partial solutions. GA-dA has the fastest improvement rate and the highest mean score on both knapsack problems. This may be the type of problem for which it is best suited, as the dA has no ordering constraints, can capture dependencies between large groups of variables and, importantly, search around these solutions. The neural network-based techniques appear to be promising methods for these constrained problems, greatly outperforming BOA and PBIL.

GA-dA is also the fastest consistent solver of the 128-bit Royal Road Problem. This problem is famous because it was used to identify the hitch-hiking phenomenon that can adversely affect the performance of GAs. For an EDA to perform well on this problem, it has to build a model that learns that either the partitions are independent or that every variable in the string is independent. PBIL suffers on this problem by falling into a local optima where certain variables have a very low probability of outputting a one. At this point it becomes very unlikely that certain subproblems will be solved. At the point of solving, BOA has learnt a model where every variable is independent. We will look more closely at the models produced by GA-dA and GA-NADE in the next section but GA-NADE clearly learns a complex relationship between variables that does not exist in the real problem. GA-dA hones in on a very small part of the sub-space, which allows it to quickly find the optimal solution.
%
\begin{figure}[t!]
\centering
\subfigure[Covariances between sub-partitions on the modified royal road]{
\makebox[1.0\textwidth]{
        \includegraphics[scale=0.35]{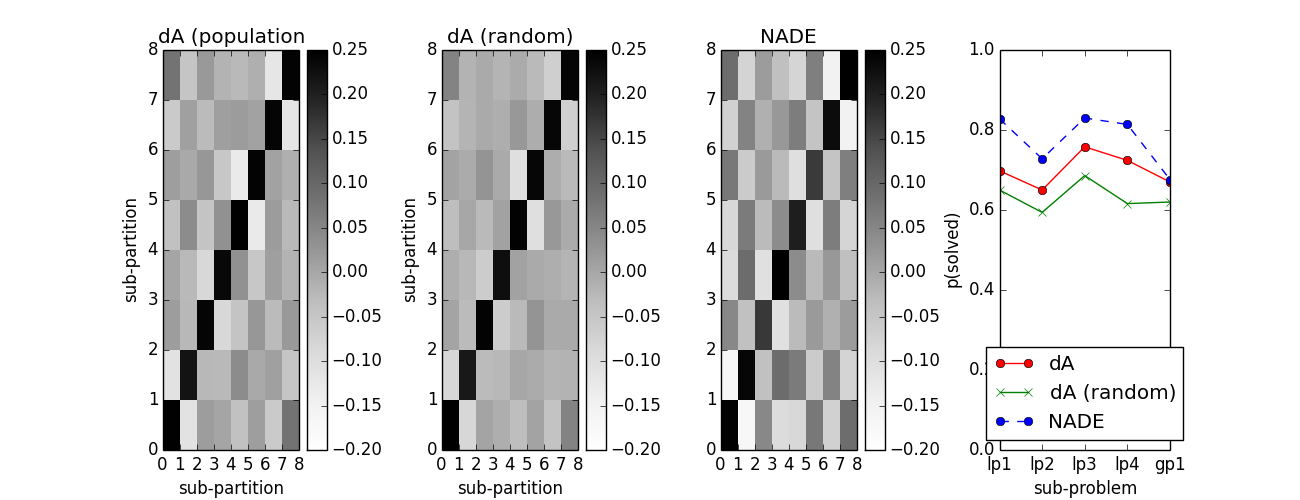}
    }
}
\subfigure[Covariances between sub-partitions on the modified royal road when the first sub-partition is fixed to all ones]{
\makebox[1.0\textwidth]{
        \includegraphics[scale=0.35]{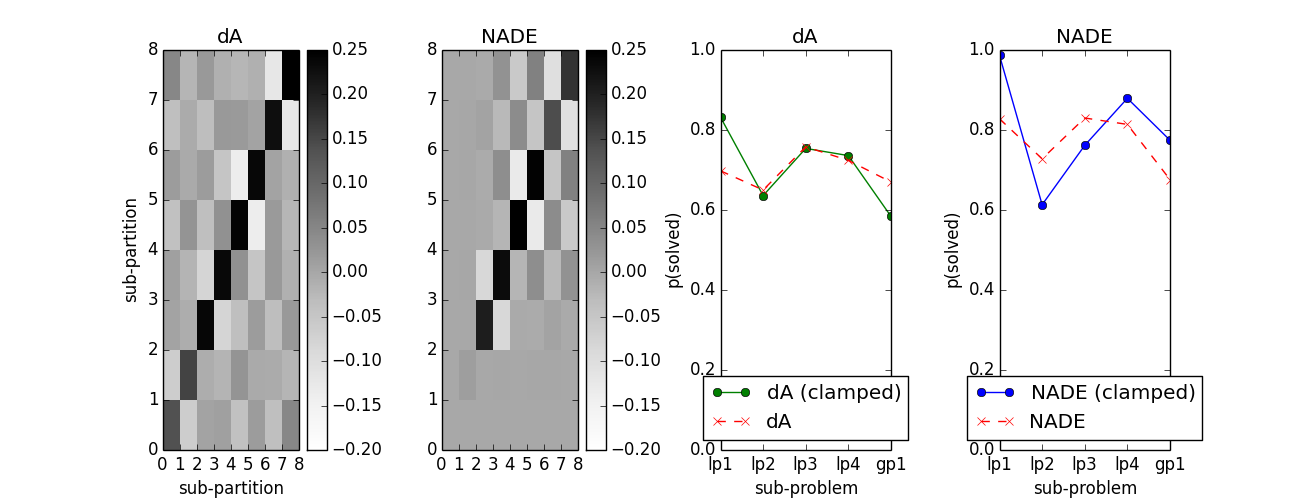}
    }
}
\subfigure[Covariances between sub-partitions on the modified royal road when the last sub-partition is fixed to all ones]{
\makebox[1.0\textwidth]{
        \includegraphics[scale=0.35]{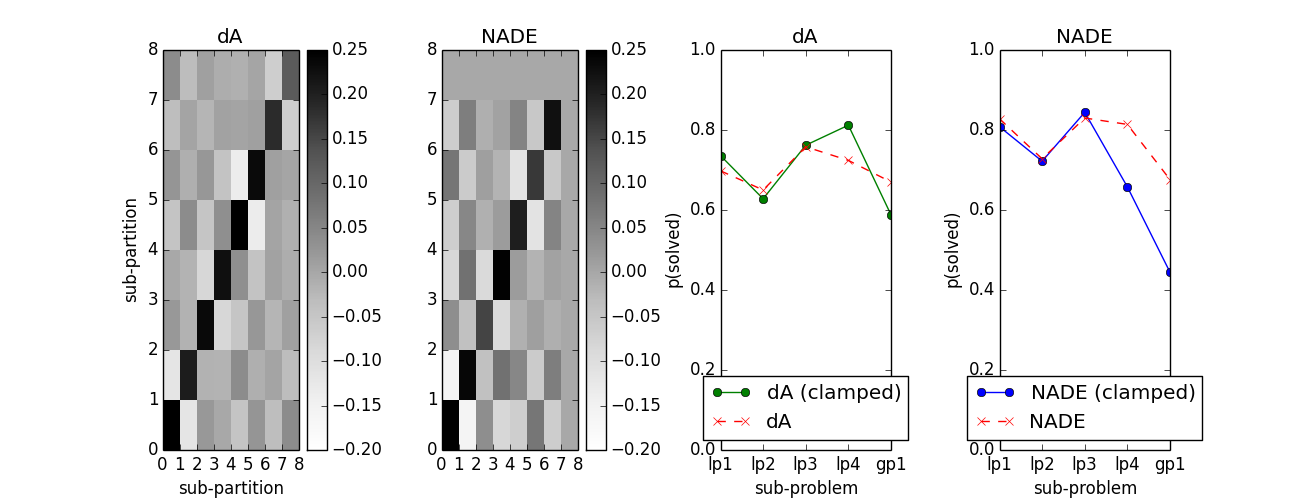}
    }
}

\caption[Optional caption for list of figures]{Showing samples from the dA and NADE on the Royal Road with Linkages problem.}
  \label{fig:nade_vs_da1}
\end{figure}

\subsection{Neural Network Methods Comparison}

There are clear differences between the performance of the two neural network-based algorithms, with GA-NADE performing much better than GA-dA on the HIFF and MaxSat problems but significantly worse on the others. The two methods discover structure in solution space and produce new solutions in considerably different ways. GA-NADE produces a generative model in the form of a Bayesian network and GA-dA through a compressed encoding of promising solutions. With GA-NADE new individuals are produced by sampling from the NADE, with no direct input from the population. With GA-dA, individuals from the population are fed forward through the dA - there is a direct relationship between the population and new solutions. In this section we first investigate the differences between the NADE and dA models on a specially designed function and then provide further analysis on a selection of the problems from the results section above.

\subsubsection{A Royal Road with Linkages}
\begin{figure}[t!]
\centering
\subfigure[Showing how a bit string is partitioned on the Royal Road with Linkages]{
\makebox[1.0\textwidth]{
        \includegraphics[scale=0.7]{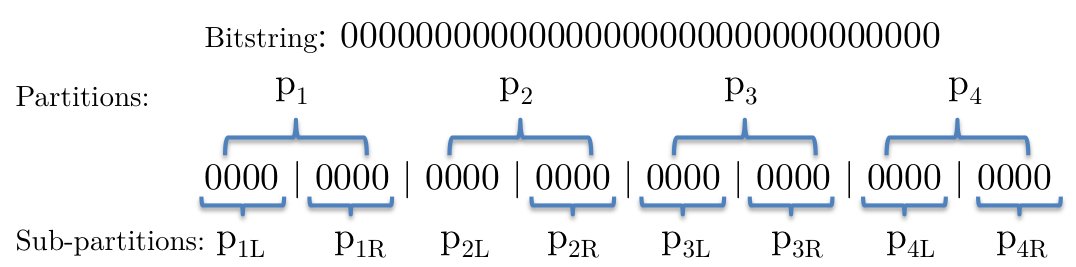}
    }
}
\subfigure[An example optimal solution on the Royal Road with Linkages]{
\makebox[1.0\textwidth]{
        \includegraphics[scale=0.7]{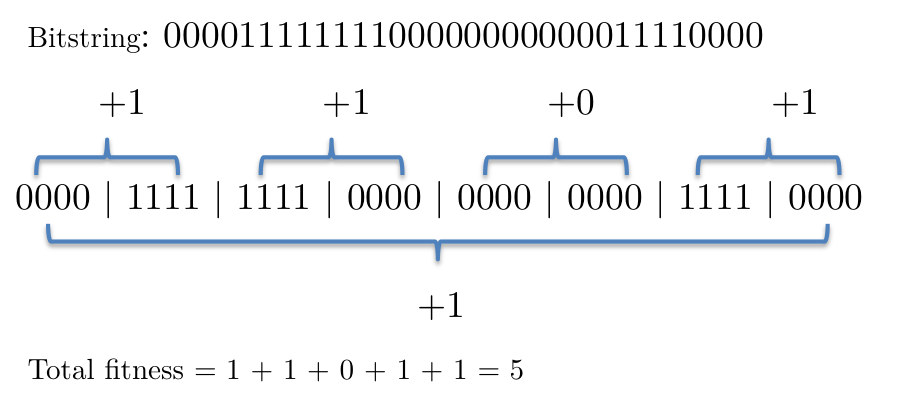}
    }
}
\caption[Optional caption for list of figures]{Showing samples from the dA and NADE on the Royal Road with Linkages problem.}
  \label{fig:churchillroadexample}
\end{figure}

To investigate the dependencies learnt by the two models, an objective function has been constructed based on the Royal Road \citep{mitchell1992royal}. A bit string is divided into equal sized partitions, where the fitness contribution of a partition depends on the state of other sub-partitions.

A solution consists of \(n\) non-overlapping partitions, \(p_i \quad i\in{[1\ ... \ n]}\), of size \(2k\). \(p_i\) is further divided into two non-overlapping partitions of size \(k\), \(\hat{p}_{iL}\) on the left side and \(\hat{p}_{iR}\) on the right. If each bit in \(\hat{p}_{iL}\) is equal to 1 and each bit in \(\hat{p}_{iR}\) is equal to 0, 1 is contributed to the total fitness. Alternatively, a 1 is contributed to the total fitness if each bit in \(\hat{p}_{iL}\) is equal to 0 and each bit in \(\hat{p}_{iR}\) is equal to 1. Finally, an additional 1 is contributed to fitness if every bit in \(\hat{p}_{1L}\) and \(\hat{p}_{nR}\) are equal to 1, or if every bit in \(\hat{p}_{1L}\) and \(\hat{p}_{nR}\) is equal to 0, where \(i=1\) signifies the first partition and \(i=n\) the last. An example is given for a 32-bit sequence in figure \ref{fig:churchillroadexample}. This fitness function has been invented to explore how the algorithms deal with dependency between blocks of variables. There is linkage between the first and second halves of each partition, which is referred to below as `local-linkage', and between the first \(k\) and the last \(k\) bits in the string, referred to as the `global-linkage'.

GA-NADE and GA-dA are applied to a 32-bit version of this problem (\(k=4,n=4\)), both using a population size of 500. Samples produced by the models at the point that the optimal solution is found is explored in figure ~\ref{fig:nade_vs_da1}. This figure shows the covariance between sub-partitions and the probability that a sub-problem is solved in a sample, given three different conditions.

Figure ~\ref{fig:nade_vs_da1}(a) compares the dA using the final population as input (dA-pop), which is the normal behaviour, the dA using input from a uniform random distribution (dA-rand) and the NADE. We first analyse the covariance matrix produced from the samples to infer whether the models have correctly learnt the relationships between variables in this solution space. All of the models display negative covariance between the local-linkage sub-partitions, which means that there is an anti-correlation between the two partitions, and positive covariance between the global sub-problem, which affects the first and last sub-partition. This shows that the models have learnt structure in this search space.

Changing the input to the dA from a uniform random distribution to individuals from the population has a small but significant effect on the covariances, strengthening the local and global relationships. Consequently there is a greater probability of solving all of the sub-problems when using the final population as input. The negative covariance between sub-partitions in the local-linkage sub-problems is strongest with the NADE, although the positive covariance between the first and last sub-partitions is similar for both models, meaning that the NADE has a significantly higher probability of solving the local-linkage sub-problems, but both have an equal probability of solving the global-linkage problem.

In Figure ~\ref{fig:nade_vs_da1}(b), we probe the models further by ''clamping" the output of the first sub-partition so that all four bits equal one, to see if this influences the second sub-partition. The first four bits of the samples inputted to the dA are clamped in a similar fashion. The probability of solving the first sub-problem is increased for both models, showing that the clamping has biased the values of the second sub-partition. The greatest effect is seen with the NADE, which now solves the first sub-problem almost 100\% of the time. The NADE also has a corresponding increase in the probability of solving the global sub-problem, which in the clamped case is decreased with the dA. 

An alternate clamping is presented in Figure ~\ref{fig:nade_vs_da1}(c). Here, the last four bits in the sequence are fixed to equal one. To gain a fitness contribution, the second to last group of 4 bits must be a vector of zeroes (the last local sub-problem) and the first four bits must be a vector of ones (global sub-problem). GA-NADE has a large decrease in the probability of solving these two sub-problems, while the probabilities of solving the other sub-problems are unchanged. The dA has an increased probability of solving the last local sub-problem, with a slight decrease in solutions to the global sub-problem. As with a Bayesian graph, the NADE has fixed dependency ordering, from left to right. Clamping the last four bits in the sequence has no influence on the preceding partitions. However, with the dA two-way dependencies can be learnt, and here it can be seen that the last sub-partition influences the preceding sub-partition. As with the first example, clamping reduces the probability of solving the global dependency, suggesting that the dA does not capture this section of the problem structure as well as the NADE.

Figure ~\ref{fig:nade_vs_da1} provide an interesting insight into the differences between the two models. Both models are able to learn linkages between several variables. Figures ~\ref{fig:nade_vs_da1}(b) and \ref{fig:nade_vs_da1}(c) demonstrates that the NADE is able to capture the global linkages much better than the dA. However, the dA does not suffer from ordering constraints, while the NADE does. The a priori choice of ordering could have a large influence on the performance of the NADE, which is a topic for further investigation. As there are no ordering constraints in the dA, this provides flexibility in biasing the variables, i.e. through clamping, which could be utilised to further guide the search process.
\begin{figure}[t!]
\centering
\subfigure[Showing the distribution of the mean distances from their 5 nearest neighbours and the fitnesses of 50,000 samples from the dA and NADE at the point that the optimal solution is found]{
\makebox[1.0\textwidth]{
        \includegraphics[scale=0.4]{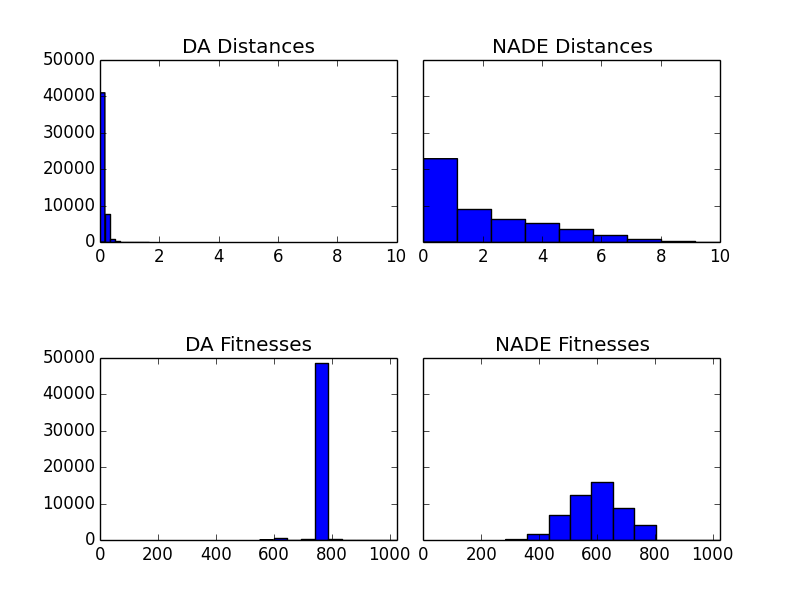}
    }
}
\subfigure[Showing 2,000 random samples (first row) and the probability of a 1 at each locus averaged over 50,000 samples (second row).]{
\makebox[1.0\textwidth]{
        \includegraphics[scale=0.4]{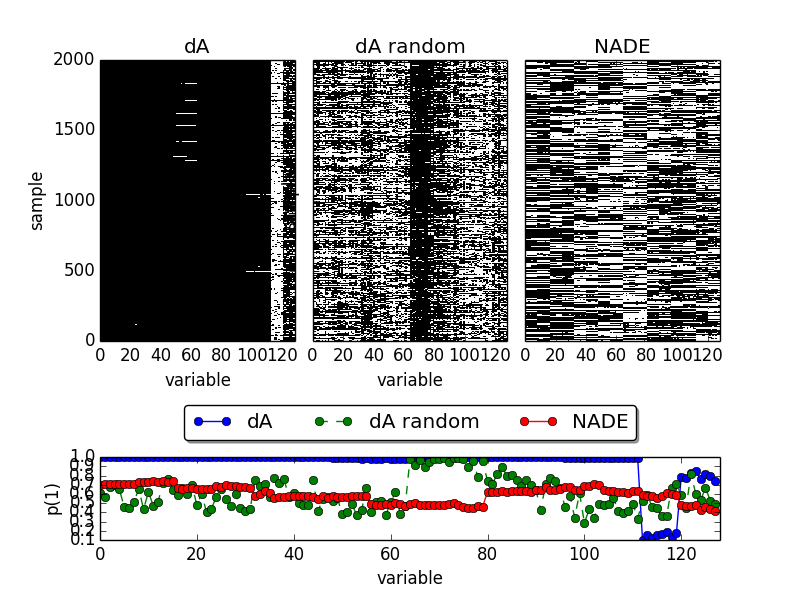}
    }
}
  \caption{Showing statistics from samples from the dA and NADE models on the 128-bit HIFF.}
  \label{fig:results_hists1}
\end{figure}

\begin{figure}[t!]
\centering
\subfigure[Showing the distribution of the mean distances from their 5 nearest neighbours and the fitnesses of 50,000 samples from the dA and NADE at the point that the optimal solution is found]{
\makebox[1.0\textwidth]{
        \includegraphics[scale=0.4]{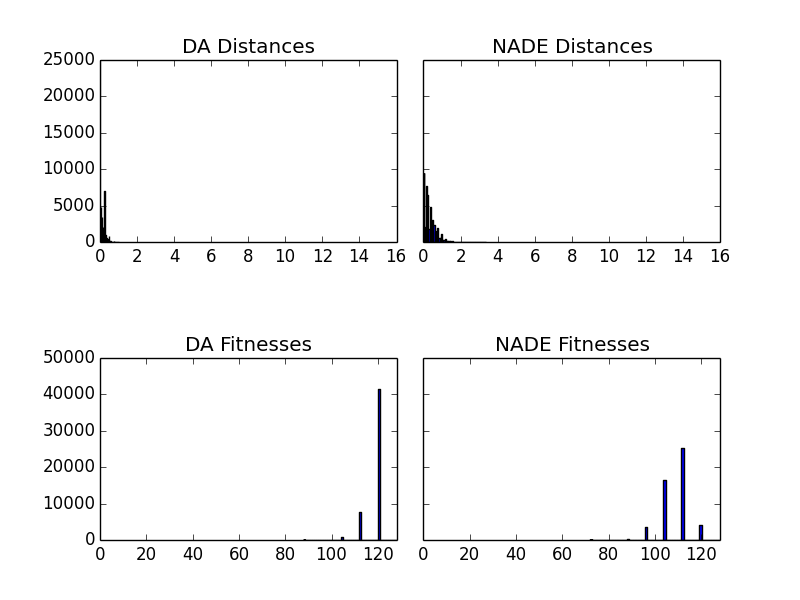}
    }
}
\subfigure[Showing 2,000 random samples (first row) and the probability of a 1 at each locus averaged over 50,000 samples (second row).]{
\makebox[1.0\textwidth]{
        \includegraphics[scale=0.4]{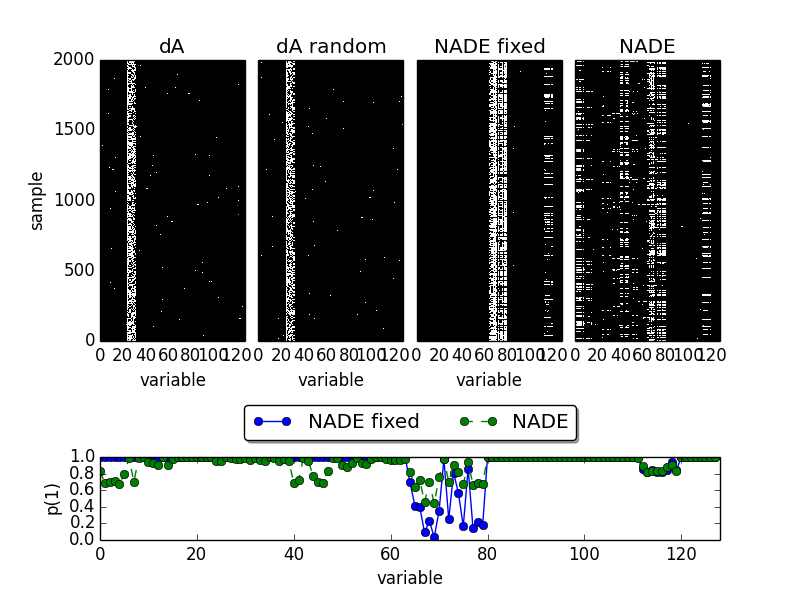}
    }
}
  \caption{Showing statistics from samples from the dA and NADE models on the Royal Road.}
  \label{fig:results_hists2}
\end{figure}

\begin{figure}[t!]
\centering
\subfigure[Showing the distribution of the mean distances from their 5 nearest neighbours and the fitnesses of 50,000 samples from the dA and NADE at the point that the optimal solution is found]{
\makebox[1.0\textwidth]{
        \includegraphics[scale=0.4]{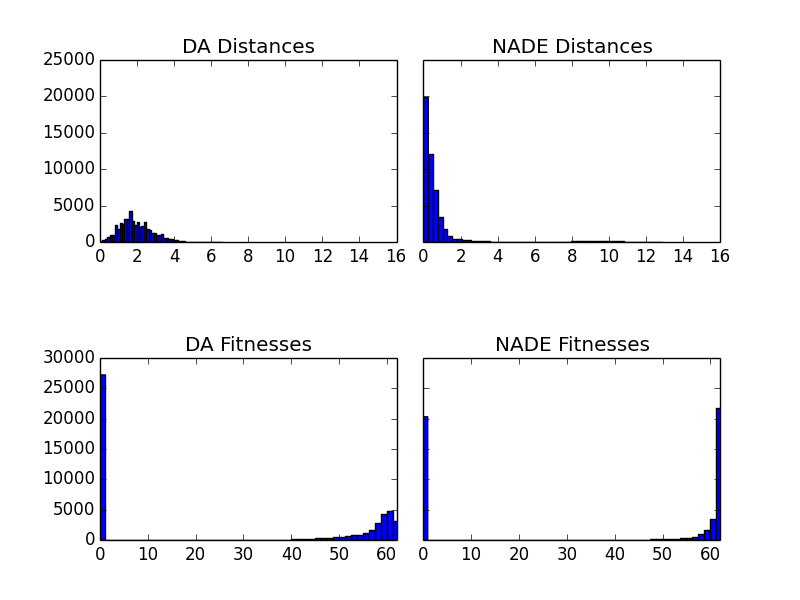}
    }
}
\subfigure[Showing 2,000 random samples (first row) and the probability of a 1 at each locus averaged over 50,000 samples (second row).]{
\makebox[1.0\textwidth]{
        \includegraphics[scale=0.4]{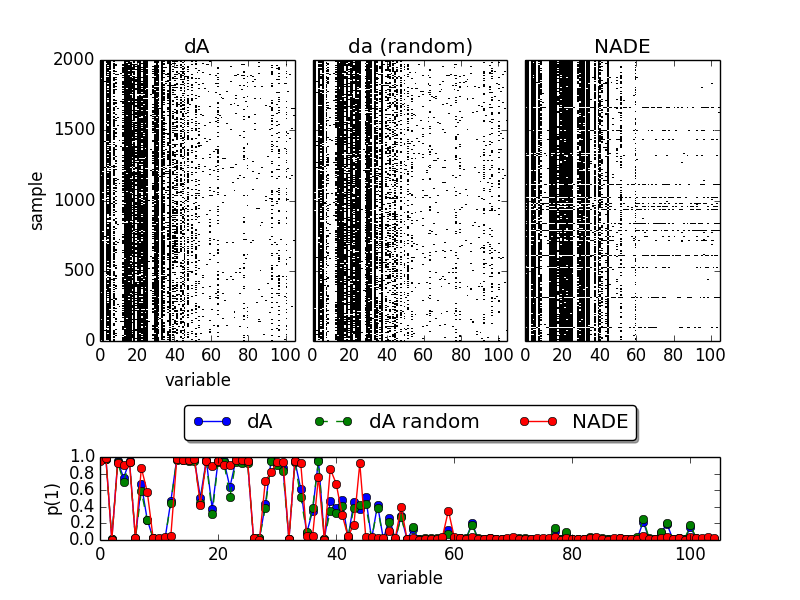}
    }
}
  \caption{Showing statistics from samples from the dA and NADE models on the Weing8 Knapsack instance.}
  \label{fig:results_hists3}
\end{figure}

\subsubsection{Tested Problems}

There is a marked difference between GA-dA and GA-NADE on several test problems, and in this section we further analyse the algorithms on the HIFF-128, the Royal Road and Weing8 problems. 50,000 samples are taken from the dA and NADE at the point that the optimal solution is found on HIFF-128. Figure \ref{fig:results_hists1}(a) shows the distribution of the mean distances of samples from their 5 nearest neighbours and the fitnesses of the samples. There is much greater diversity in the samples produced by the NADE, while the dA has a much more peaked distribution. This can be partly explained by the fact that the dA does not use a niching strategy on this problem (a choice based on the results of a grid search). On this problem, most of the dA's solutions are small mutations around the local optima of 784, while the NADE produces a much more diverse set of samples. Looking at a selection of samples in Figure \ref{fig:results_hists1}(b), we see that the dA model using the population as input is extremely peaked, with almost all variables in the string outputting a one apart from the last 16 bits. Using input sampled from a uniform random distribution produces a more varied distribution of solutions, although the fitness of the samples is not as high. The NADE produces a less biased distribution of samples, which explains its better performance on this problem. Of particular interest is that samples from the NADE contain partial solutions from both the `all ones' and `all zeroes' global optima, while the dA concentrates on the `all ones' optima. This implies that NADE is better suited towards capturing the distribution of multi-modal search spaces.

Figure \ref{fig:results_hists2}(a) shows a similar distribution of samples is produced by the dA on the Royal Road problem, with 15 of the 16 8-bit partitions outputting ones and mutations around a single unfilled partition. Although the NADE produces more diverse samples, the distribution is peaked with a very low probability of a one from bits 64 - 80 (partitions 8 and 9). Figure \ref{fig:results_hists2}(b), the `NADE fixed' plot shows the samples produced when the outputs of the first 64 bits are clamped to a one. This increases the probability of producing a zero in the bits 64 - 80, suggesting that a false dependency has been learnt linking earlier parts of the bit string to partitions 15 and 16, making it very unlikely that the optimal solution will occur in a sample. This could be due to the topology of the dependency graph that the NADE is tied to, i.e. a node on the left is connected to every node on the right. For the dA model, we see in this case that there is very little difference between using input sampled from a random distribution or from the population, which is due to the high level of corruption (\(p(c)=0.9\)).

For the Weing8 knapsack problem instance, the histograms in Figure \ref{fig:results_hists3}(a) show that the NADE produces slightly less diversity in its samples. The fitnesses of the samples are also heavily skewed towards very high scoring. Both models produce a large number of infeasible (and thus low scoring) samples. The raster plots in Figure \ref{fig:results_hists3}(a) indicate that both models concentrate on similar solutions, with a peaked response to many variables on the left side of the string (bits 50 and under). As indicated in the aggregate plot in Figure \ref{fig:results_hists3}(b), the NADE has a more peaked response to many variables than the dA. We see in the raster plot that there are a large number of samples from the NADE that use fewer items from the left side of the string and more from the right side (leading to the appearance of lines in the plot). This suggests that, similarly to the HIFF example, the NADE can model a multi-modal distribution, which could be useful for multi-modal search spaces or multi-objective optimisation. The corruption level is relatively high in the dA in this example (\(p(c)=0.25\)), and there is only small differences between using the population and a random sample as input to the model. However, sampling from the population increases the probability of certain variables expressing a 1 in a number of positions.

The analysis in this section has shown that there are differences between the two neural network models. In the HIFF and Royal Road examples, the dA converges to a very peaked distribution, producing samples which are similar, while the NADE produces much more diversity. This is particularly apparent in the HIFF, where the NADE produces solutions close to the `all ones' and `all zeroes' global optimums, suggesting that the NADE is better at capturing a multi-modal distribution. However, the NADE can find spurious linkage between variables in the search space, which can make it extremely difficult to find the optimal solution. 

\subsection{Areas for Further Investigation}
In this paper we have demonstrated how the dA and NADE models can be successfully incorporated into an evolutionary search. Both models outperform a GA on number of problems (the dA is never beaten by a GA) as well as PBIL and BOA on certain problems. There are a number of areas that are currently being investigated that could potentially improve performance. Both models are trained online, and we have seen in the previous section that a bias can creep into the models. For example, in the Royal Road problem the NADE has a strong bias towards certain partitions being set to zeroes, keeping the search in a local optima for many evaluations. Training online allows search space structure discovered earlier in the search process to be reused later. It also helps keep down the computational cost of model building. However, rebuilding the models either at certain predefined intervals, at every iteration or when probabilities begin to peak could reduce sample bias and improve the performance of the algorithm. As an aside, there could be another benefit to training online, which is the potential for transfer of the models between different related instances of the same class of problem. The question of transfer learning - can building a model to solve one task help improve performance on another? - has not been answered by the EDA community, although recent work has made a start (e.g. \cite{hauschild2012using}). Preliminary investigations suggest that using a dA, information learned in one task can be retained and recombined with that learnt in a second task to produce an across the board speed up on the third task in the sequence. Careful experimentation is needed to discover which types of problem can use machine learning techniques to transfer knowledge between unique instances and speed up optimisation as more examples are encountered.

In the last decade there has been a resurgence of interest in neural network-based methods, spurred by so called `deep architectures'. Both the dA and NADE could potentially benefit from the hierarchical features supported by deep networks and this needs to be investigated further. A potential problem with the NADE model is the precedence ordering of variables, which could have a great effect on the quality of the learnt distribution, as well as the training time. There are several paths that could be taken in this direction, for example having multiple models being learnt in parallel or using alternate methods to learn Bayesian networks at certain points in the search process and using the discovered ordering for the NADE model.

With both models, although especially with the NADE, the addition of local search could improve performance by providing more exploitation and fine-tuning of solutions, as seen in \citep{pelikan2003hierarchical}. This would allow the dA or NADE to power a more global search and a share of the responsibility for exploitation to the local search procedure. In the same vein, an interesting extension of the algorithms would be to hybridise them with a selecto-recombinative evolutionary algorithm (EA). Here, the EA could maintain the population of solutions, recombining and mutating them, while the neural network models are trained on the population. The EA could draw upon the models to generate genotypes, to guide search in a direction learnt from the structure of the solution space. 
\subsection{Real Valued, Mixed Integer and Multi-objective Problems}
Both the NADE and dA can be modified to work with continuous or mixed-integer domains, providing an advantage over other discrete only EDA methods. As briefly mentioned earlier, the outputs of the dA can be interpreted as parameters of a distribution from which new solutions are sampled. Therefore, for real-valued problems, the outputs of the dA can be interpreted as means of a Gaussian distribution. Similarly, instead of predicting the means of a Bernouilli distribution, the NADE can be set up to predict the means and variances of a one-dimensional mixture of Gaussians. This real-valued variant of the NADE is called the RNADE (\cite{uria2013rnade}). It is also be possible to mix Bernouilli and Guassian output units, to optimise Mixed Integer Problems, an underrepresented area in EDA optimisation. The fact that the proposed algorithms can be generalised to real-valued problems is very useful since other popular EDAs like BOA and hBOA work only for discrete-valued data.

\section{Conclusion}
We have presented a novel stochastic evolutionary algorithm based on two neural network models that iteratively learn to produce more effective genotypes. Training online, using the best found genotypes, the models are able to adaptively learn an exploration distribution, which guides search towards optimal solutions on difficult problems such as the 256-bit HIFF and a 430 3-CNF MaxSat problem, regularly outperforming a Genetic Algorithm and other EDA methods on a number of problems. 
\small

\bibliographystyle{apalike}
\bibliography{ecjsample}

\end{document}